\documentclass[10pt,journal,compsoc]{IEEEtran}

\ifCLASSOPTIONcompsoc
  \usepackage[nocompress]{cite}
\else
  \usepackage{cite}
\fi

\ifCLASSINFOpdf
\else
\fi
\usepackage{algorithm,algorithmic,amsmath,amssymb,array,cite,color,colortbl,comment,eqparbox,float,graphics,graphicx,subfigure,times,url,xcolor}

\begin{document}

\title{SenTion: A framework for Sensing Facial Expressions}

\author{
    \IEEEauthorblockN{Rahul Islam\IEEEauthorrefmark{1}, Karan Ahuja\IEEEauthorrefmark{1}, Sandip Karmakar\IEEEauthorrefmark{1}, Ferdous Barbhuiya\IEEEauthorrefmark{1}}
    \IEEEauthorblockA{\IEEEauthorrefmark{1}IIIT Guwahati
    \\\{rahul.islam, karan.ahuja, sandip, ferdous\}@iiitg.ac.in}
}

\IEEEtitleabstractindextext{%
\begin{abstract}
Facial expressions are an integral part of human cognition and communication, and can be applied in various real life applications. A vital precursor to accurate expression recognition is feature extraction. 
In this paper, we propose SenTion: A framework for sensing facial expressions. We propose a novel person independent and scale invariant method of extracting Inter Vector Angles (IVA) as geometric features, which proves to be robust and reliable across databases. SenTion employs a novel framework of combining geometric (IVA's) and appearance based features (Histogram of Gradients) to create a hybrid model, that achieves state of the art recognition accuracy. We evaluate the performance of SenTion on two famous face expression data set, namely: CK+ and JAFFE; and subsequently evaluate the viability of facial expression systems by a user study.  Extensive  experiments showed that SenTion framework yielded dramatic improvements in facial expression recognition and could be employed in real-world applications with low resolution imaging and minimal computational resources in real-time, achieving 15-18 fps on a 2.4 GHz CPU with no GPU.   
\end{abstract}

}

\maketitle

\IEEEdisplaynontitleabstractindextext
\IEEEpeerreviewmaketitle

\section{Introduction}
\label{sec:introduction}

Facial expressions play a very important role for sensing the perceived human emotion and intention in people. Driven by advancements in machine learning, image processing and human cognition they have found a wide variety of applications in interpreting user reactions, human computer interaction systems, emotion based interactive games, creation of virtual characters, surveillance, driver-less car, etc.

Facial expression analysis consist of three major stages: feature extraction, feature selection and classifier construction. In the first stage, features based on geometry or appearance textures of the face are extracted from a image or from a sequence of images. These facial features are extracted from face by using various feature extraction method like HOG, LBP, Gabor-Wavelet, AMM models etc. From these extracted set feature a subset feature is selected in second stage which best fitted to distinguish between different expression. Then in third stage classifier is constructed using these subset of features.
Although various methods has been proposed for facial expression recognition, robust recognition that performs well across databases and subjects is still a challenging task, as most of the methods are person dependent, intolerant  to scale changes and have low rate of accuracy across emotions and high computational overhead. \par 
In this paper, we propose SenTion: a person independent and scale invariant framework for sensing emotions from facial expressions. Accuracy of the proposed system has been tested on the Cohn-Kanade Plus and JAFFE database, where it shows considerable improvement with low computational overhead making is suitable for realtime application. We validate our results with a user study, that compares the recognition accuracy of our system across different expressions and overlays it with the corresponding human perception associated with it. The specific contributions of our work is threefold. We first describe in detail our novel techniques of fusing geometrical and appearance based features. We then demonstrate the goodness of our system by evaluating it on benchmark databases. Finally, we validate the practicality of facial expression recognition systems via a first of it's kind user study that measures the ease of expressing and understanding of various facial expressions by humans.   
\vspace*{-5mm}
\section{Related Work}
\label{sec:relwork}
Among the various communications mechanism human being uses such as speech, hand gestures and emotions etc, emotions and the encoded feelings prove to be an important factor for an effective communication. One easy of way of understanding emotion is to analyze facial expression which have significance in human communication, interaction and can be also applied in facial biometric-based intelligent systems.\par

Among the notable contribution for determining facial expression Yacoob et al.\cite{de2000probabilistic} fitted local parametric motion models on face and then classified the parameters by a nearest neighbor classier to recognize expression. Hoey et al. \cite{hoey2001hierarchical} used a multilevel Bayesian network to learn in a weakly supervised manner the dynamics of facial expression. De la Torre et al. \cite{de2007temporal} proposed a geometric-invariant clustering algorithm to decompose a stream of one person's facial behavior into facial gestures. Shan et al. \cite{shan2009facial} explored the use of local binary patterns (LBP) for the given task. Bartlett et al. \cite{bartlett2005recognizing} proposed a method based on gabor filters and AdaSVMs for the same purpose.\par

For facial expression detection mainly two types of feature are used: appearance based and geometry based. Appearance features describe the texture of the face caused by expression, such as wrinkles and furrows. Geometric features describe the shape of the face and its components such as the mouth or the eyebrows \cite{valstar2012meta}. Among the appearance based feature, Gabor wavelet representation is widely adopted in facial expression analysis \cite{bartlett2005recognizing} \cite{1613024}. These gabor features are largely affected by scaling and computation of gabor features is time intensive. Local Binary Patterns(LBP) is also an effective appearance features for facial expression analysis \cite{zhao2007dynamic} \cite{happy2012real} \cite{shan2009facial} . Geometric feature based approaches use active appearance models (AAMs) or similar kind technique to track a dense set of facial points on face. The locations of these points are then used to infer the shape of facial features such as the mouth or the eyebrows and thus to classify the facial expression \cite{valstar2012meta}. Sebe et al. \cite{sebe2007authentic} proposed a geometric feature based system where he used Piecewise Bézier volume deformation tracking after manually locating a number of facial points.

Recently researcher have employed deep learning for the problem of facial expression analysis. Liu et al. \cite{liu2013aware} used hybrid deep learning model i.e. CNN + RBM for automatic detection basic facial expression. Kim et al. \cite{kim2013deep} compared different models of Deep Belief Network(DBN) for unsupervised feature learning in audio-visual emotion recognition. Liu et al. \cite{liu2014facial} proposed a Boosted Deep Belief Network(BDBN) to perform three task namely feature learning, feature selection and classifier construction on CK+ and JAFFE Dataset. In contrast to these computationally intensive models, our work aims to integrate the simplicity of geometric models and the thoroughness for salient feature extraction of appearance based models, by combining the two to form a hybrid model which achieves state of the art accuracy while being robust and reliable across databases.\par 
\vspace*{-2mm}
\section{Model Considerations}
\label{sec:model}

\begin{figure}[thb]
\includegraphics[width=0.5\textwidth]{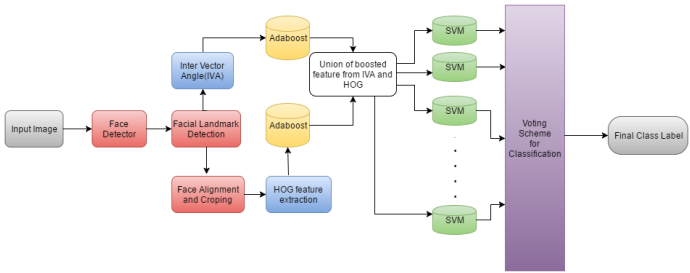}
\caption{Schematic representation of proposed model.}
\label{fig:overview}
\end{figure}

In this section, we present the primary modules of our proposed facial expression recognition algorithm. Figure~\ref{fig:overview} illustrates our proposed framework of feature extraction, training and subsequent testing. \par

It has been observed that facial expressions are usually manifested by contraction and expansion of facial muscles which alters the position of facial landmarks. Along with facial muscles the texture of the area also changes. \cite{happy2015automatic}
Our proposed work attempts to exploit the changes by making use of geometric features to capture changes in facial landmarks and appearance based features for capturing more salient features such as texture and contour.

For each image, the area of the face is localized first, followed by the detection of various facial landmarks. We adapt the concept of scale invariant Inter Vector Angles (IVA) to act as geometric features, which relies upon prior facial landmarks for feature computation. We use this approach in junction with appearance based features, which are represented by Histogram of Gradients proposed by \cite{felzenszwalb2010object}. Prior to HOG feature extraction, we crop the face from the image using face detection and then align and normalize the Region of Interest for uniform extraction and training of features across different datasets. Once the feature vectors are obtained, they are passed through AdaBoost for feature selection and then Support Vector Machines for learning and classification. 

\subsection{Facial Landmark Detection}

Happy et al. \cite{happy2015automatic} states that the importance of accurate facial land marking is undeniable to achieve better performance for facial expression recognition. A pre-requisite to facial landmark extraction is face detection, for which we make use of the dlib library\cite{King}, which provides a powerful frontal face detector that is robust to pose changes. We then detect 68 facial landmarks from the above detected image using Vahid et al \cite{Kazemi_2014_CVPR} state of the art real time face pose estimation algorithm. Their algorithm uses a general framework based on gradient boosting for learning an ensemble of regression trees for estimating the face's landmark positions directly from a sparse subset of pixel intensities. 

\vspace*{-4mm}
\subsection{Geometry Based Feature}
\label{geometry}
Many existing literature's such as \cite{baltrusaitis2015cross} use the 2D locations of facial landmark in the image space as geometric features. Similar angle based representation have also been explored by  \cite{tian2001recognizing}, \cite{valstar2007distinguish} and \cite{cohn2014automated}. However, their approaches are only defined in the discrete image space and lack a mathematical formulation. Hence, while such models can adapt to different scales, they fail to capture the individual differences of faces. 
To overcome these shortcomings, we propose the concept of Inter Vector Angle(IVA). In this work, we not only study the holistic spatial analysis of geometric features, but derive a novel mathematical formulation of angle based representations. The generality of our foundation helps it to span over various feature landmarks, without being custom designed for a specific feature region such as lips, eyes, brow, etc. as is the case in \cite{tian2001recognizing}.
IVA's are scale invariant geometric features computed from the previously extracted facial landmark as follows:

Let P be the set of the location of all 2D landmarks such that $(P_1, P_2, P_3, ... P_{68}) \in P$ where $P_i$ denoted the (x,y) coordinate of the facial landmark in image space where i=1:68. Let $V_{ij}$ denote the vector between any two facial landmark $P_i$ and $P_j$ such that $i \neq j$. Therefore $\overrightarrow{V_{ij}}=P_i - P_j$. The set of all such vector $\overrightarrow{V}$ can be denoted by $\overrightarrow{V} \Leftarrow \forall \overrightarrow{V_{ij}}$, where $i \neq j$. To compute the we select any two vectors $\overrightarrow{V_{ij}}, \overrightarrow{V_{mn}} \in \overrightarrow{V}$ such that $i = j$ and $m \neq n$. Now the angle between any two such vector is given by:

\begin{equation}
\label{eq:angle}
\theta_{ijn} = \cos^{-1} (\frac{\overrightarrow{V_{ij}}. \overrightarrow{V_{mn}}}{|\overrightarrow{V_{ij}}| |\overrightarrow{V_{mn}}|})
\end{equation}

where . denotes the dot product and $|\overrightarrow{x}|$ signifies the magnitude of the vector $\overrightarrow{x}$. We define $\theta \Leftarrow \forall \theta_{ijn}$ such that $j \neq n$. Here $\theta$ denotes the set of all angles between any two vectors joining a set of facial landmarks, and having a common point. 

We observe that the cardinality of $\theta$ to be 150348. This is because we have $|P|=68$. The total number of triangle that can be formed, has the maximum possibility of $^{68}C_3$ considering no three points are co-linear. Each triangle has three angles therefore leading to $|\theta|=3 \times {^{68}C_3}=150348$. Out of these, the points which are co-linear do not contribute to feature weightage, and their angles are assigned to zero.

The use of $\theta$ as IVA geometric feature imbibes the scale invariant property of angles in our system. Hence, the face Region of Interest no longer needs to be taken into normalized space leading to lossy down-sampling as is the case with most appearance and geometric features which depend upon the size of the image such as those implemented in state of the art algorithms such as in  \cite{happy2015automatic} and \cite{zhang2011facial}. We also observe in section \ref{IVA_good} that IVA out performs traditional geometric features such as those which consider only the vector length or landmark locations.

\subsection{Face Alignment}

In order to make our feature model scale space and rotation invariant, we need to map it to a common reference frame. This also helps us to better capture and analyze the texture of the face. We use the eye centers (averaged from the detected eye corner) to act as the reference points when computing the similarity transform to obtain a cropped, scaled and rotated face. The resultant is a 192X192 pixel image of the face. 

\vspace*{-1mm}
\subsection{Appearance Based Features}
\label{appearance}
Appearance based feature are extracted on aligned facial images. We extract Histogram of Gradient(HOGs) of aligned face based on the configuration proposed by Felzenswalb et al.\cite{felzenszwalb2010object}, prior to which we use Gaussian Pyramids to upscale the image\cite{bradski2008learning}. HOG features are defined using 9 contrast insensitive gradient orientations, a cell size of 8 and truncation 0:2. This leads to a 36-dimensional feature vector. We have used Scikit-learn\cite{Pedregosa:2011:SML:1953048.2078195} implementation of HOG to extract HOG features. 

\vspace*{-1mm}
\subsection{Feature Boosting}
\label{boost}
Literature suggest that Adaboost and SVM works in harmony \cite{garcia2007boosting}\cite{li2008adaboost}\cite{li2005study} \cite{1613024}. Using SVM as a weak learner for Adaboost, it performs not only as a fast classifier, but also as a feature selection technique. An advantage of feature selection by Adaboost is that features are selected contingent on the features that have already been selected\cite{ghimire2013geometric}. 
In our work, we use the Decision Tree Classifier as the base estimator and use SAMME.R as the boosting algorithm \cite{zhu2009multi}. We select the best feature obtained by applying Adaboost for every class using One-Against-All strategy. We do this by binning the best weak-learners obtained in all the iterations for every one-against-all sub-problem and picking the top 'k' features. The final feature set is taken to be the union of best features obtained for every sub-problem.

\subsection{Classification and Validation}
\label{classification}

For face expression detection, we use Support Vector Machine(SVM) as our learning model. SVM\cite{cortes1995support} is a popular machine learning algorithm which maps the feature vector to a different plane, usually to a higher dimensional plane, by a non-linear mapping, and finds a linear decision hyper plane for classification of two classes. Since SVM is binary classifier multiclass problem can be extended using two methods: One-Against-One (OAO) or One-Against-All (OAA).\par
In our work, we have used OAO method to detect multiple expression using SVM. SVMs are trained for each combination of classes and the final class label is decided using majority voting strategy. 
Hence, ${^{K}C_2}$ number of classifiers need to be constructed where K is the number of classes (facial expressions to be classified).
\section{Evaluation}
\label{sec:evaluation}

In this section, we discuss the performance of our methodology. For experiments, we use a hardware configuration of Intel Pentium CPU 2020M @ 2.40 GHz and 4 GB RAM. We run our experiments on the two widely used facial expression datasets, that is, Japanese Female Facial Expression (JAFFE) \cite{lyons1998coding} and Cohn-Kanade (CK+)\cite{ck}. We use 10 fold cross validation to evaluate the performance of our proposed methodology. We train our method using both geometric (IVA) and appearance based (HOG) features, and construct ${^{6}C_2}$ number of SVM classifiers for evaluating the performance on the test set.

\subsection{Experiments on the Cohn-Kanade Database}
The Cohn-Kanade+ database consists of 593 posed sequences from 123 subjects, with all images having dimensions 640x490 pixels. Each sequence in the database goes from neutral to target display, with the last frame being AU encoded. 
The images contain six different facial expressions: anger, disgust, fear, happiness, sadness and surprise.\cite{zhong2012learning} For our experiments, we use the images whose sequences could be labeled as one of these six basic facial expressions.
We showcase the performance of different feature selection techniques, namely: Inter Vector Angles (IVA), Histogram of Gradients (HOG) and a hybrid model using HOG + IVA. We use AdaBoost for feature selection as the ensemble method, employing a Decision Tree Classifier as the base estimator and maximum number of estimators as 100. This generates a feature vector of size 531 and 508 for IVA and HOG respectively, giving us a total value of k (as described in section ~\ref{boost}) as 1039.   



\subsubsection{Using Geometric Features: }
\label{IVA_good}
We test the accuracy of our system using Geometric Features, namely Inter Vector Angles described in section ~\ref{geometry} with AdaBoost described in section ~\ref{boost}. As this method is scale invariant we do not need to re-size the image, hence making it applicable on the fly. A critique of most geometric feature and shape based models is that the distance between facial landmarks vary from person to person, thereby making the system person dependent and less reliable \cite{happy2015automatic}. However, the use of IVA's make the features independent of people as they are found to be reliable across subjects. If we choose only vector lengths (i.e. distance between landmarks) as features, we achieve a limited accuracy of 87.9\%, as compared to the current system of IVA features which achieves an accuracy of 97.42\%. This establishes the robustness of our system.\par
\begin{figure}[tbh]
\includegraphics[width=0.45\textwidth]{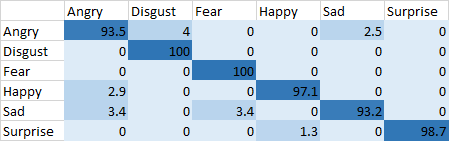}
\caption{Confusion Matrix of IVA on CK+ database}
\label{fig:CM_IVA_CK}
\vspace*{-3mm}
\end{figure}

  
As compared to traditional methods that make use of Gabor Wavelets, such as those proposed by \cite{bartlett2005recognizing} and \cite{1613024} are affected by the resolution of the image, our method is more computationally sensitive and is scale invariant.The method also beats the state of the art accuracies reported by appearance based models such as the 94.09\% accuracy of \cite{happy2015automatic}, which employ appearance based salient feature patches and 93.14\% of \cite{zhang2011facial} which employ both appearance based features (in the form of salient patches) and geometric features (in the form of salient distances) in their method using Facial Movement Features. The proposed system achieves an average F1 Score of 96.8\%, precision of 97.29\% and recall of 97.12\%. Figure ~\ref{fig:CM_IVA_CK} depicts the confusion matrix of this method.  

\subsubsection{Using Appearance Features: }
\begin{figure}[tbh]
\includegraphics[width=0.45\textwidth]{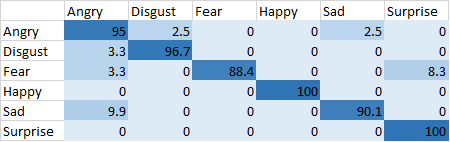}
\caption{Confusion Matrix of HOG on CK+ database}
\label{fig:CM_HOG_CK}
\end{figure}

Histogram of Gradients (HOGs) are widely used for automatic facial action unit detection and have showcased state of the art performance in the same. \cite{baltrusaitis2015cross}. 
We make use of HOGs with AdaBoost as proposed in section ~\ref{appearance} and ~\ref{boost} respectively to test it's individual performance and contribution to the system. Using the HOG features extracted from the aligned face images, we achieve an accuracy of 96.73\%, precision of 96.87\%, recall of 95\% and F1 Score of 95.42\%. This depicts that HOG are very robust at capturing the texture and salient patches of the face, without prior knowledge of discriminative patches. Figure ~\ref{fig:CM_HOG_CK} depicts the confusion matrix of this method. 

\subsubsection{Hybrid Model: IVA + HOG: } 
\begin{figure}[tbh]
\centering
\includegraphics[width=0.45\textwidth]{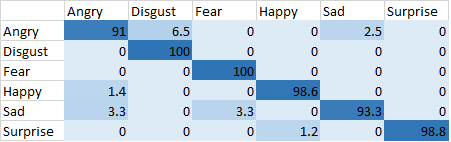}
\caption{Confusion Matrix of Hybrid Model on CK+ database}
\label{fig:CM_IVA_HOG_CK}
\end{figure}

We combine the scale invariant property of Inter Vector Angles with the texture and salient patches capturing property of Histogram of Gradients, to create a hybrid model which is an amalgam of both geometric and appearance based features. This model, achieved an accuracy of 97.38\%, with a precision of 97.24\%, recall of 96.94\% and F1 Score of 96.71\%. The confusion matrix for the same can be found in Figure ~\ref{fig:CM_IVA_HOG_CK}. Table ~\ref{tab:performance_ck} compares the performance of this method from the results obtained by other approaches reported in literature. Subsequently Table ~\ref{tab:performance_rank_ck} showcases the ranks of these methods. In both the tables we compare the accuracy across 6 facial expressions: Angry, Disgust, Fear, Happy, Sad and Surprise. We attain the best rank out of all the methods, with an average rank of 1.66. Our approach is the best for classifying images belonging to Anger, Fear and Disgust. Due to the simplicity of computation, scale in-variance and robustness of our model, it can be used in real life applications and on devices with constrained resources such as tablets and mobiles. Our model is stable, efficient and does not require any GPU's for training and even outperforms current Deep Learning techniques in terms of accuracy on the CK+ database. Our method beats the average accuracy of 92.22\% achieved by Liu et al.\cite{liu2013aware} using AU-Aware Deep Networks(AUDN's) and 99.6\% achieved by \cite{burkert2015dexpression} which employs deep convolutional neural networks. It also exceeds the 8 fold cross validation accuracy of 96.7\% achieved by \cite{liu2014facial} using Boosted Deep Belief Networks (BDBN). While other deep neural network architectures such as \cite{khorrami2015deep} and \cite{jeni2014spatio} exist, they predict Facial Action Units rather than categorizing facial expressions and hence our approach cannot be compared to their's.
\par 
Therefore, being able to perform significantly better on the standard CK+ database with low computational requirement, high speed of 15 frames per second and stability, makes our work the most applicable among the known state of the art.     

\begin{table}[bhtp]
\begin{center}
\begin{tabular}{|c|l|c|c|c|c|c|c|}
\hline
\bf{Method} & \bf{Anger} & \bf{Disgust} & \bf{Fear} & \bf{Happy} & \bf{Sad} & \bf{Surprise} \\
\hline
Uddin\cite{uddin2009enhanced} & 82.5 & 97.5 & 95 & 100 & 92.5 & 92.5 \\
\hline
Poursaberi\cite{poursaberi2012gauss} & 87.03 & 91.58 & 90.98 & 96.92 & 84.58 & 91.23 \\
\hline
Zhong\cite{zhong2012learning} & 71.39 & 95.33 & 81.11 & 95.42 & 88.01 & 98.27 \\
\hline
Song\cite{song2010image} & 90.56 & 86.04 & 84.61 & 93.61 & 90.24 & 92.3 \\
\hline
Zhang\cite{zhang2011facial} & 87.1 & 90.2 & 92 & 98.07 & 91.47 & 100 \\
\hline
Liu\cite{liu2013aware} & 81.48 & 95.48 & 82.67 & 99.52 & 71.43 & 97.59 \\
\hline
Happy\cite{happy2015automatic} & 87.8 & 93.33 & 94.33 & 94.2 & 96.42 & 98.46 \\
\hline
\color{blue} Our Method & \color{blue}91 & \color{blue}100 & \color{blue}100 & \color{blue}98.5 & \color{blue}93.2 & \color{blue}98.7 \\
\hline
\end{tabular}
\end{center}
\caption{Performance comparison (in percentage) of different Facial Expression Recognition methods on the CK+ Database}
\label{tab:performance_ck}
\vspace*{-5mm}
\end{table}

\begin{table}[bhtp]
\begin{center}
\begin{tabular}{|l|c|c|c|c|c|c|c|}
\hline
\bf{Method} & \bf{An} & \bf{Di} & \bf{Fe} & \bf{Ha} & \bf{Sa} & \bf{Su} & \bf{Avg. Rank}\\
\hline
\cite{uddin2009enhanced} & 6 & 2 & 2 & 1 & 3 & 6 & 3.33 \\
\hline
\cite{poursaberi2012gauss} & 5 & 6 & 5 & 5 & 7 & 8 & 6 \\
\hline
\cite{zhong2012learning} & 8 & 4 & 8 & 6 & 6 & 4 & 6 \\
\hline
\cite{song2010image} & 2 & 8 & 6 & 8 & 5 & 7 & 6 \\
\hline
\cite{zhang2011facial} & 4 & 7 & 4 & 4 & 4 & 1 & 4 \\
\hline
\cite{liu2013aware} & 7 & 3 & 7 & 2 & 8 & 5 & 5.33 \\
\hline
\cite{happy2015automatic} & 3 & 5 & 3 & 7 & 1 & 3 & 3.66 \\
\hline
\color{blue} Our's & \color{blue}1 & \color{blue}1 & \color{blue}1 & \color{blue}3 & \color{blue}2 & \color{blue}2 & \color{blue}1.66 \\
\hline
\end{tabular}
\end{center}
\caption{Performance-based ranks and average ranks of State-of-the art methods on the CK+ Database. A lower value indicates a better rank.}
\label{tab:performance_rank_ck}
\vspace*{-7mm}
\end{table}

\subsection{Experiments on JAFFE Database}
The JAFFE database contains 213 gray images of seven facial expressions (six basic + neutral) taken from 10 Japanese Female models, with each image having a resolution of 256X256 pixels. In this paper, all the images of six basic expressions from the JAFFE database are used, namely: anger, disgust, fear, sad, happy and surprise. 

\begin{figure}[bhtp]
\includegraphics[width=0.5\textwidth]{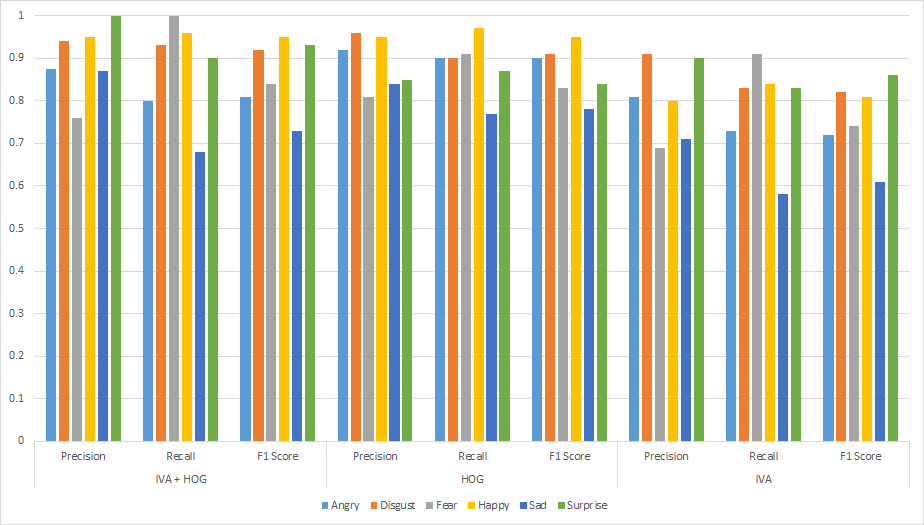}
\caption{Performance of our method on the JAFFE database for different facial expressions.}
\label{fig:performance_metrics_jaffe}
\end{figure}
\begin{figure}[bhtp]
	\subfigure[Inter Vector Angles]{
		\includegraphics[width=0.4\textwidth]{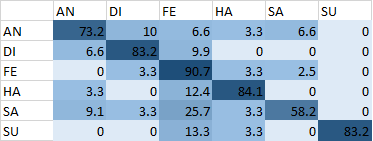}
		\label{fig:CM_IVA}
	}
	\subfigure[Histogram of Gradients]{
		\includegraphics[width=0.4\textwidth]{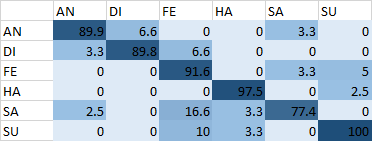}
		\label{fig:CM_HOG}
	}
	\subfigure[Hybrid Model: IVA + HOG]{
		\includegraphics[width=0.4\textwidth]{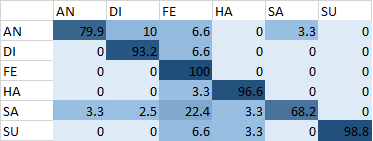}
		\label{fig:CM_Hybrid}
	}
\caption{Confusion matrix of proposed methods on the JAFFE database. Here: AN $\rightarrow$ Anger, DI $\rightarrow$ Disgust, FE $\rightarrow$ Fear, HA $\rightarrow$ Happy, SA $\rightarrow$ Sad, SU $\rightarrow$ Surprise}
\label{fig:CM_JAFFE}
\vspace{-0.2in}
\end{figure}

To evaluate the robustness and accuracy of our system on different databases, we test the accuracy of our system on the JAFFE database. While testing on the JAFFE database, we use the same parameters as those used in the CK+ database; so as to evaluate the reliability of our system with images of different face sizes and resolutions without additional adaptation. We found the JAFFE database to be more challenging than the CK+ database and hence set the maximum number of estimators in AdaBoost to 400, to avoid loss of important features. Figure ~\ref{fig:performance_metrics_jaffe} denotes the  performance (F1 score, precision and recall) of our method on the various facial expressions. 
We test our model using both geometric (IVA), appearance based (HOG) and a hybrid model of features as proposed in section ~\ref{sec:model}. The confusion matrix for the same can be found in Figure ~\ref{fig:CM_JAFFE}. It is interesting to note that using IVA, we obtain an accuracy of 78.89\% and using HOG we obtain an accuracy of 88.83\%. Using our hybrid model (IVA+HOG) we get an accuracy of 88.2\%. As opposed to the previous case (in CK+ dataset), appearance based features have a greater impact than geometric features on the JAFFE dataset. Our proposed hybrid model correctly captures this, and remains stable and adapts well to the changes across databases. 

\subsection{User Study}

In order to find test the feasibility of facial expression systems for various applications such as those described in section ~\ref{sec:introduction}, we perform a user study. Here we ask 5 people to showcase various near-frontal facial expressions as depicted in Table ~\ref{tab:user_study}. It has to be kept in mind that these people were not trained actors/actresses and were not provided any guidance on how to showcase their expressions. They were then asked to give their rating on the ease of expressing the respective facial expressions on the well established Likert scale\cite{likert}, on a scale of 1 to 5. Subsequently, the ease of understanding their facial expressions was measured by a set of 3 human evaluators by means of a blind testing, wherein each evaluator was showed a 5 second clip of the respective expression and was asked to identify the facial expression and then rate it on a Likert Scale form 1 to 5. A miss-classification of expression was rated 0. Every evaluator measured the ease of interpreting each facial expression of each person. The average ratings of the user study are depicted in Table ~\ref{tab:user_study}. According to the table; happy, surprise and angry are the top 3 facial expressions that can be expressed and understood easily by humans. Also, there is a direct correlation between ease of expression and ease of understanding a facial expression, both being directly proportional to one another.

\begin{table}[bhtp]
\begin{center}
\begin{tabular}{|c|c|c|c|c|c|c|c|}
\hline
\bf{Emotion} & \bf{Ease of Expression} & \bf{Ease of Understanding} \\
\hline
\bf{Happy} & 4.41 & 4.20\\
\hline
\bf{Surprise} & 3.78 & 3.08\\
\hline
\bf{Sad} & 3.1 & 2.02\\
\hline
\bf{Angry} & 3.3 & 2.76\\
\hline
\bf{Disgust} & 2.6 & 2.42\\
\hline
\bf{Fear} & 2.1 & 0.665\\
\hline
\end{tabular}
\end{center}
\caption{Overall ratings of the user Study}
\label{tab:user_study}
\vspace*{-10mm}
\end{table}

\section{Discussion}
\label{sec:discussion}

We have trained and cross validated the performance of our system on both CK+ and JAFFE dataset. The accuracy of 97.38\% (hybrid model) on the CK+ dataset, shows the proposed approach is robust and out performs other state of the art methods.   

Analyzing the confusion matrix of our hybrid model on the CK+ dataset in Figure ~\ref{fig:CM_IVA_HOG_CK} and JAFFE dataset in Figure ~\ref{fig:CM_Hybrid}, we find that  
the proposed method sometimes confuses anger with sadness and disgust. This is due to the similar geometric and appearance features between the facial expressions, especially in the JAFFE database, where the difference is less stark. Other than this, no major problem could be detected.

From the experiment on CK+ dataset we have seen that IVA features gives more accuracy than combined HOG and IVA features (by 0.04\%). In JAFFE dataset we have seen that HOG features gives more accuracy than combined HOG and IVA (by 0.63\%). From this we can can conclude that IVA geometric features more dominant in CK+ and HOG appearance features are more dominant in JAFFE database. Therefore a system comprising both feature sets (IVA + HOG) proves to be more stable, reliable and accurate to perform facial expression recognition across databases without additional adaptation. 

The user study and it's subjective evaluation also showcases the viability of facial expressions acting as a precursor for emotions in an induced environment. While a claim can be made that person-dependent and naturally occurring facial expressions can provide a better insight, the study presents vital information such as which facial expression to use for practical applications in an induced setting where the system cues the user for depicting an emotion such as that for human-computer interaction or creation of virtual characters. Therefore, while the expressions depicted in the user study lacked peak intensity it provided a good sample of facial expressions in everyday life.
\vspace*{-1mm}
\section{Conclusion and Future Work}
\label{sec:conclusion}

This paper has presented a computationally efficient facial expression recognition system (SenTion) for accurate and robust classification of
the six universal expressions. We proposed a novel approach of taking Inter Vector Angles (IVA) as geometric features, which proved to be scale invariant and person independent. It's combination with appearance based Histogram of Gradients, provided a hybrid model which was reliable and stable across databases without additional adaptation. The effectiveness of the proposed approach of IVA's and appearance HOG based features is validated by it's performance on the CK+ and JAFFE database and comparison with other state-of-the art methods. We outperformed the state of the art algorithm on the CK+ database, achieving an accuracy of 97.38\%. \par 
The proposed approach being simplistic, consistent, computationally sensitive and real time can be potentially applied into many applications, such as human emotional state detection, patient monitoring, fatigue detection and tutoring systems. In our future work, we will extend our approach to in the wild settings and also test it's performance on video based facial expression recognition systems.

\bibliographystyle{IEEEtran}
\bibliography{egbib}

\end{document}